\documentclass[10pt,twocolumn,letterpaper]{article}

\usepackage{cvpr}
\usepackage{times}
\usepackage{epsfig}
\usepackage{graphicx}
\usepackage{amsmath}
\usepackage{amssymb}
\usepackage{subfigure}
\graphicspath{ {figs/} }
\usepackage{booktabs}       
\usepackage{float}
\usepackage{array}
\usepackage{caption}
\usepackage{multirow}

\usepackage{color}

\usepackage{amsmath,amssymb}
\include{figs}

\usepackage[breaklinks=true,bookmarks=false]{hyperref}

\cvprfinalcopy 

\def\cvprPaperID{****} 

\newcolumntype{L}[1]{>{\raggedright\let\newline\\\arraybackslash\hspace{0pt}}m{#1}}
\newcolumntype{C}[1]{>{\centering\let\newline\\\arraybackslash\hspace{0pt}}m{#1}}
\newcolumntype{R}[1]{>{\raggedleft\let\newline\\\arraybackslash\hspace{0pt}}m{#1}}
\begin{document}
\title{On the iterative refinement of densely connected representation levels for semantic segmentation}
\author{Arantxa Casanova$^{1,2}\ $ Guillem Cucurull$^{1,2}\ $ 
Michal Drozdzal$^{1,3}\ $ Adriana Romero$^{1,3}\ $
Yoshua Bengio$^1$\\
$^1$ Montreal Institute for Learning Algorithms $\quad$
$^2$ Computer Vision Center, Barcelona \\
$^3$ Facebook AI Research \\
}

\maketitle

\begin{abstract}
State-of-the-art semantic segmentation approaches increase the receptive field of their models by using either a downsampling path composed of poolings/strided convolutions or successive dilated convolutions. However, it is not clear which operation leads to best results. In this paper, we systematically study the differences introduced by distinct receptive field enlargement methods and their impact on the performance of a novel architecture, called Fully Convolutional DenseResNet (FC-DRN). FC-DRN has a densely connected backbone composed of residual networks. Following standard image segmentation architectures, receptive field enlargement operations that change the representation level are interleaved among residual networks. This allows the model to exploit the benefits of both residual and dense connectivity patterns, namely: gradient flow, iterative refinement of representations, multi-scale feature combination and deep supervision. In order to highlight the potential of our model, we test it on the challenging CamVid urban scene understanding benchmark and make the following observations: 1) downsampling operations outperform dilations when the model is trained from scratch, 2) dilations are useful during the finetuning step of the model, 3) coarser representations require less refinement steps, and 4) ResNets (by model construction) are good regularizers, since they can reduce the model capacity when needed. Finally, we compare our architecture to alternative methods and report state-of-the-art result on the Camvid dataset, with at least twice fewer parameters.
\end{abstract}
\vspace{-0.1in}
\section{Introduction}
Convolutional Neural Networks (CNNs) have been extensively studied in the computer vision literature to tackle a variety of tasks, such as image classification \cite{He2015,Huang2016,Hu2017}, object detection \cite{He2017maskrcnn} and semantic segmentation \cite{Jegou17,ChenPSA17}. Major advances have been driven by novel \emph{very deep} architectural designs \cite{He2015,Huang2016}, introducing skip connections to facilitate the forward propagation of relevant information to the top of the network, and provide shortcuts for gradient flow. Very deep architectures such as residual networks (ResNets) \cite{He2015}, densely connected networks (DenseNets) \cite{Huang2016} and squeeze-and-excitation networks \cite{Hu2017} have exhibited outstanding performance on standard large scale computer vision benchmarks such as ImageNet \cite{ILSVRC15} and MSCOCO \cite{mscoco}.

Among top performing classification networks, ResNets challenge the hierarchical representation learning view of CNNs \cite{LiaoP16,Veit2016,Greff2016}. The hierarchical representation view associates the layers in network to different levels of abstraction. However, contrary to previous architectures such as \cite{Simonyan2014}, dropping or permuting almost any layer in a ResNet has shown to only minimally affect their overall performance \cite{Veit2016}, suggesting that the operations applied by a single layer are only a small modification to the identity operation. Significant effort has been devoted to analyzing and understanding these findings. On one hand, it has been argued that ResNets behave as an ensemble of shallow networks, averaging exponentially many subnetworks, which use different subsets of layers \cite{Veit2016}. On the other hand, it has been suggested that ResNets engage in an unrolled iterative estimation of representations, that refine upon their input \cite{Greff2016}. These arguments have been exploited in \cite{Drozdzal2017} to learn normalized inputs for iterative estimation, highlighting the importance of having transformations prior to the residual blocks.

Fully Convolutional Networks (FCNs) were presented in \cite{Long2015,Ronneberger2015} as an extension of CNNs to address per pixel prediction problems, by endowing standard CNNs with an upsampling path to recover the input spatial resolution at their output. In the recent years, FCN counterparts and enhanced versions of top performing classification networks have been successfully introduced in the semantic segmentation literature. Fully Convolutional ResNets (FC-ResNets) were presented and analyzed in \cite{Drozdzal2016} in the context of medical image segmentation. Moreover, Fully Convolutional DenseNets (FC-DenseNets) \cite{Jegou17} were proposed to build low capacity networks for semantic segmentation, taking advantage of iterative concatenation of features maps.

In this paper, we further exploit the iterative refinement properties of ResNets to build densely connected residual networks for semantic segmentation, which we call Fully Convolutional DenseResNets (FC-DRNs). Contrary to FC-DenseNets \cite{Jegou17}, where the convolution layers are densely connected, FC-DRN apply dense connectivity to ResNets models. Thus, our model performs iterative refinement at each representation level (in a single ResNet) and uses dense connectivity to obtain refined multi-scale feature representations (from multiple ResNets) in the pre-softmax layer. We demonstrate the potential of our architecture on the challenging CamVid \cite{camvid} urban scene understanding benchmark and report state-of-the-art results. To compare and contrast with common pipelines based on top performing classification CNNs, we perform an in depth analysis on different downsampling operations used in the context of semantic segmentation: dilated convolution, strided convolution and pooling. Although dilated convolutions have been well adopted in the semantic segmentation literature, we show that such operations seem to be beneficial only when used to finetune a pre-trained network that applies downsampling operations (e.g. pooling or strided convolution). When trained from scratch, dilation-based models are outperformed by their pooling and strided convolutions-based counterparts, highlighting the generalization capabilities of downsampling operations.

The contributions of our paper can be summarized as:
\begin{itemize}
\itemsep0em 
\item We combine FC-DenseNets and FC-ResNets into a single model (FC-DRN) that fuses the benefits of both architectures: gradient flow and iterative refinement from FC-ResNets as well as multi-scale feature representation and deep supervision from FC-DenseNets.
\item We show that FC-DRN model achieves state-of-the-art performance on CamVid dataset \cite{camvid}. Moreover, FC-DRN outperform FC-DenseNets, while keeping the number of trainable parameters small.
\item We provide an analysis on different operations enlarging the receptive field of a network, namely poolings, strided and dilated convolutions. We inspect FC-DRN by dropping ResNets from trained models as well as by visualizing the norms of the weights of different layers. Our experiments suggest that \emph{the benefits of dilated convolutions only apply when combined with pre-trained networks that contain downsampling operations}. Moreover, we show that \emph{ResNets (by model construction) are good regularizers, since they can reduce the model capacity at different representation levels when needed, and adapt the refinement steps.}
\end{itemize}

\section{Related work}
In the recent years, FCNs have become the \emph{de facto} standard for semantic segmentation. Top performing classification networks have been successfully extended to perform semantic segmentation \cite{Visin2015,Quan2016,Drozdzal2016,Wu2016,Jegou17}. 

In order to overcome the spatial resolution loss induced by successive downsampling operations of classification networks, several alternatives have been introduced in the literature; the most popular ones being long skip connections in encoder-decoder architectures~\cite{Long2015,SegNet2015, Ronneberger2015,islam2017gated} and dilated convolutions~\cite{Yu2015,Chen2014}. Long skip connections help recover the spatial information by merging features skipped from different resolutions on the contracting path, whereas dilated convolutions enlarge the receptive field without downsizing the feature maps.

Another line of research seeks to endow segmentation pipelines with the ability to enforce structure consistency to their outputs. The contributions in this direction include Conditional Random Fields (CRFs) and its variants (which remain a popular choice) \cite{Koltun11,Chen2014,Zheng2015}, CRFs as Recurrent Neural Networks \cite{Zheng2015}, iterative inference denoising autoencoders \cite{Romero2017}, convolutional pseudo-priors \cite{Xie2016}, as well as graph-cuts, watersheds and spatio-temporal regularization~\cite{Beier2016,Quan2016,kundu2016feature}. 

Alternative solutions to improve the performance of segmentation models are based on combining features at different levels of abstraction. Efforts in this direction include iterative concatenation of feature maps \cite{Huang2016,Jegou17}; fusing upsampled feature maps with different receptive fields prior to the softmax classifier~\cite{Chen2016}, along the network \cite{Lin2017,Ardiyanto2017} or by means of two interacting processing streams operating at different resolutions \cite{Pohlen2017}; gating skip connections between encoder and decoder to control the information to recover~\cite{islam2017gated}; and using a pyramid pooling module with different spatial resolutions for context aggregation~\cite{Zhao2017}. Moreover, incorporating global features has long shown to improve semantic segmentation performance~\cite{Gatta2014, Liu2015}.

Finally, semantic segmentation performance has also been improved by training with synthetic data~\cite{Richter}, propagating information through video data~\cite{Jampani2016}, or modeling uncertainties in the model~\cite{Kendall2017}.

\section{Fully Convolutional DenseResNet}
In this section, we briefly review both ResNets and DenseNets, and introduce the FC-DRN architecture.

\begin{figure*}[!htbp]
\begin{subfigure}[FC-DRN architecture]{
\includegraphics[width=0.8\textwidth]{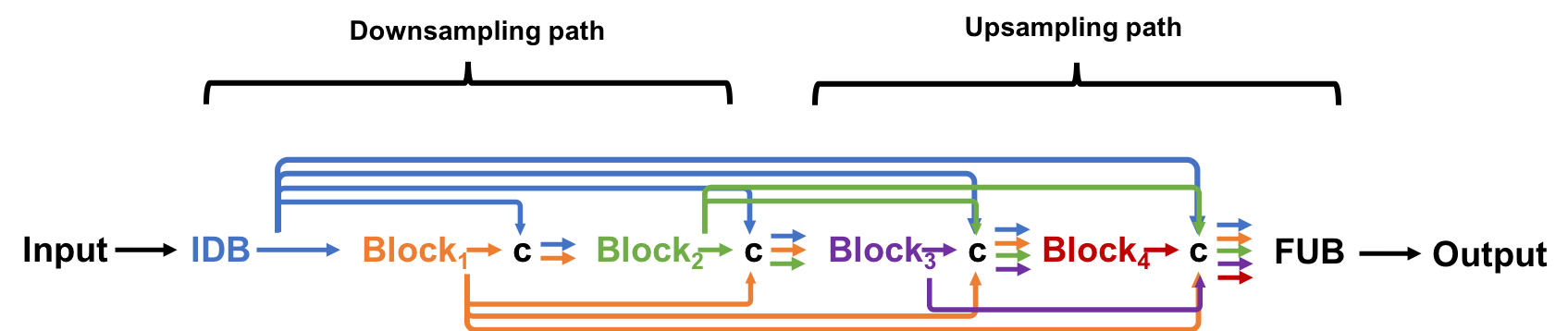}
   \label{fig:architectures_1}
}
\end{subfigure}
 \centering
\begin{subfigure}[Block]{
\includegraphics[width=0.1\textwidth]{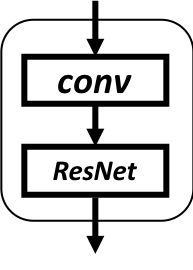}
   \label{fig:architectures_2}
}\end{subfigure}%
\caption{(a) Diagram representing our model, a dense connectivity with four Residual Networks. We use color coding to indicate the connectivity pattern in our model. IDB stands for Initial Downsampling Block and FUB stands for Final Upsampling Block. The outputs of consecutive operations are concatenated (letter c on the figure). Whenever needed, we apply transformations over the paths marked with arrows. In the downsampling path, we apply either pooling operation, strided convolution or dilated convolution to increase the network's receptive field. In the upsampling path, we apply upsampling operations to compensate for pooling or strided convolutions and $1 \times 1$ convolution for dilations. (b) Each \textit{Block} contains a convolution operation followed by a Residual Network.}
\label{fig:architectures}
\end{figure*}

\subsection{Background}
Let us denote the feature map representation of the $l$-th layer of the model as $x_l$. Traditionally, in CNNs, the feature map $x_l$ is obtained by applying a transformation $H$, composed of a convolution followed by a non-linearity, to the $l-1$-th feature map $x_{l-1}$ as $x_l = H(x_{l-1})$. CNNs are built by stacking together multiple such transformations. However, due to the non-linearity operation, optimization of such networks becomes harder with growing depth. Architectural solutions to this problem have been proposed in ResNets~\cite{He2015} and DenseNets~\cite{Huang2016}. 

In ResNets, the representation of $l$-th feature map is obtained by learning the residual transformation $H$ of the input feature map $x_{l-1}$ and summing it with the input $x_{l-1}$. Thus, the $l$-th feature map representation can be computed as follows: $x_l = H(x_{l-1}) + x_{l-1}$. This simple modification in network's connectivity introduces a path that has no non-linearities, allowing to successfully train networks that have hundreds (or thousands) of layers. Moreover, lesion studies performed on ResNets have opened the door to research directions that try to better understand how these networks work. Following these lines, it has been suggested that ResNets layers learn small modifications of their input (close to the identity operation), engaging in an iterative refinement of their input.

In DenseNets, the $l$-th feature map is obtained by applying a transformation $H$ to all previously obtained feature maps such that $x_l = H([x_0, x_1, \cdots, x_{l-1}])$, where $[\cdot, \cdot]$ denotes the concatenation operation. One can easily notice that when following the dense connectivity pattern in DenseNets, the pre-softmax layer receives the concatenation of all previous feature maps. Thus, DenseNets introduce deep feature supervision by means of their model construction. It has been shown that using the deep connectivity pattern one can train very deep models that outperform ResNets~\cite{Huang2016}. Moreover, it is worth mentioning that combining information at different representation levels has shown to be beneficial in the context of semantic segmentation \cite{Gatta2014,Jegou17}.

\subsection{FC-DRN model}
\label{ssec:fcdrn}

FC-DRNs extend the FC-DenseNets architecture of \cite{Jegou17} and incorporate a dense connectivity pattern over multiple ResNets\footnote{Note that in \cite{Jegou17} the dense connectivity pattern is over convolutional operations.}. Thus, FC-DRNs combine the benefits of both architectures: FC-DRNs perform iterative estimation at each abstraction level (by using ResNets) and combine different abstraction levels while obtaining deep supervision (by means of DenseNets connections). 

The connectivity pattern of FC-DRN is visualized in Figure \ref{fig:architectures}. First, the input is processed with an Initial Downsampling Block (IDB) composed of a single convolution followed by $2 \times 2$ pooling operation and two $3 \times 3 $ convolutions. Then, the output is fed to a \emph{dense block} (the densely connected part of the model), which is composed of ResNets, transformations and concatenations, forming a downsampling path followed by an upsampling path. 

In our model, there are 9 ResNets, motivated by the standard number of downsampling and upsampling operations in the FCN literature. Each ResNet is composed of 7 basic blocks, computing twice the following operations: batch normalization, ReLU activation, dropout and 3x3 convolution. After each ResNet, we apply a \textit{transformation} with the goal of \emph{changing the representation level}. This transformation is different in the downsampling and upsampling paths: in the downsampling path, it can be either a pooling, a strided convolution or a dilated convolution; whereas in the upsampling path, it can be either an upsampling to compensate for pooling/strided convolution or a $1 \times 1$ convolution in case of dilated convolutions, to keep models with roughly the same capacity. Following \cite{Chen2014, ChenPSA17}, transformations in the dilation-based model adopt a multi-grid pattern (for more details see Figure 1 in the supplementary material). 

The outputs of the transformations are concatenated such that the input to the subsequent ResNet incorporates information from all the previous ResNets. Concatenations are performed over channel dimensions and, if needed, the resolution of the feature maps is adjusted using transformations that are applied independently to each concatenation input\footnote{Note that in order to maintain the number of transformations when comparing different models (e.g. pooling-based vs. dilation-based model), we apply a convolution even when concatenating same resolution feature maps.}. After each concatenation, there is a $1 \times 1$ convolution to mix the features. Finally, the output of the dense block is fed to a Final Upsampling Block (FUB) that adapts the spatial resolution and the number of channels in the model output. A detailed description of the architecture is avail- able in Table 1 of the supplementary material. 

\section{Analysis and Results}
In this section, we assess the influence of applying different kinds of transformations between different ResNets and report our final results.

All experiments are conducted on the CamVid~\cite{camvid} dataset, which contains images of urban scene. Each image has a resolution of $360 \times 480$ pixels and is densely labeled with 11 semantic classes. The dataset consists of 367, 101 and 233 frames for training, validation and test, respectively. In order to compare different architectures, we report results on the validation set with two metrics: mean intersection over union (mean IoU) and global accuracy.

All networks were trained following the same procedure. The weights were initialized with HeUniform \cite{ini_he}, and the networks were trained with RMSProp optimizer \cite{rmsprop}, with a learning rate of $1e-3$ and an exponential decay of $0.995$ after each epoch. We used a weight decay of $1e-4$ and dropout rate of $0.2$. The dataset was augmented with horizontal flipping and crops of 324x324. We used early stopping on the validation mean IoU metric to stop the training, with a patience of 200 epochs.

\subsection{FC-DRN transformation variants}
State-of-the-art classification networks downsample their feature maps' resolution by successively applying pooling (or strided convolution) operations. In order to mitigate the spatial resolution loss induced by such subsampling layers, many segmentation models only allow for a number of subsampling operations and change the remaining ones by dilated convolutions \cite{Chen2014,Yu2015, Yu2017}. However, in some other cases \cite{Jegou17,noh2015learning,Ronneberger2015}, the number of downsampling operations is preserved, recovering fine grained information from via long skip connections. Therefore, we aim to analyze and compare the influence of pooling/upsampling operations versus dilated convolutions. To that aim, we build sister architectures, which have an initial downsampling block, followed by a dense block, and a final upsamping block, as described in Section \ref{ssec:fcdrn}, and only differ in the transformation operations applied within their respective dense blocks. 

\textbf{Max-Pooling architecture (FC-DRN-P):} This architecture interleaves ResNets with four max-pooling operations (downsampling path) and four nearest neighbor upsamplings followed by 3x3 convolutions to smooth the output (upsampling path).

\textbf{Strided convolution architecture (FC-DRN-S):} This architecture interleaves ResNets with four strided convolution operations (downsampling path) and four nearest neighbor upsamplings followed by 3x3 convolutions to smooth the output (upsampling path).

\textbf{Dilated architecture (FC-DRN-D):} This architecture interleaves ResNets with four multi-grid dilated convolution operations of increasing dilation factor (2, 4, 16 and 32)\footnote{We tested many different variants of dilation factors and found out that this multi-grid structure gives the best results.}. and standard convolutions to emulate the upsampling operations. Note that the dense block of this architecture does not change the resolution of its feature maps.

\textbf{FC-DRN-P finetuned with dilations (FC-DRN-P-D):} This architecture seeks to mimic state-of-the-art models based on top performing classification networks, which replace the final subsampling operations with dilated convolutions \cite{Chen2014,Yu2015, Yu2017}. More precisely, we substitute the last two pooling operations of FC-DRN-P by dilated convolutions of dilation rate 4 and 8, respectively. Following the spirit of FC-DRN-D, the first two upsampling operations become standard convolutions. We initialize our dilated convolutions to the identity, as suggested in~\cite{Yu2015}. 

\textbf{FC-DRN-S finetuned with dilations (FC-DRN-S-D):} Following FC-DRN-P-D, we substitute the last two strided convolution operations of FC-DRN-S by dilated convolutions (rates 4 \& 8), whereas the first two upsampling operations become standard convolutions. In this case, we initialize the weights of the dilated convolutions with the weights of the corresponding pre-trained strided convolutions.

Table~\ref{tab:comparison} reports the validation results for the described architectures. Among the networks trained from scratch, FC-DRN-P achieves the best performance in terms of mean IoU by a margin of 0.8\% and 3.7\% w.r.t. FC-DRN-S and FC-DRN-D, respectively. When finetuning the pooling and strided convolution architectures with dilations, we further improve the results to 81.7\% and 81.1\%, respectively. It is worth noting that we also tried training FC-DRN-P-D and FC-DRN-S-D from scratch, which yielded worse results, highlighting the benefits of pre-training with poolings/strided convolutions, which capture larger contiguous contextual information.

\begin{figure*}[!htbp]
 \centering
\begin{subfigure}[Test image]{
\includegraphics[width=0.2335\textwidth]{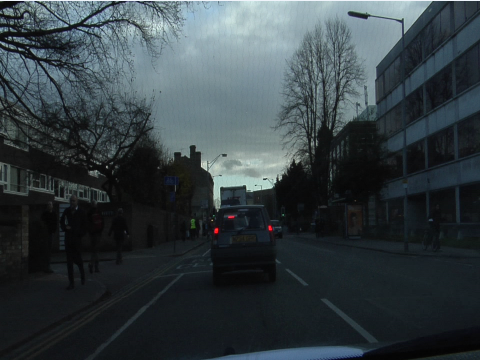}
   \label{fig:test_gt}
}
\end{subfigure}
\begin{subfigure}[Ground truth]{
\includegraphics[width=0.2335\textwidth]{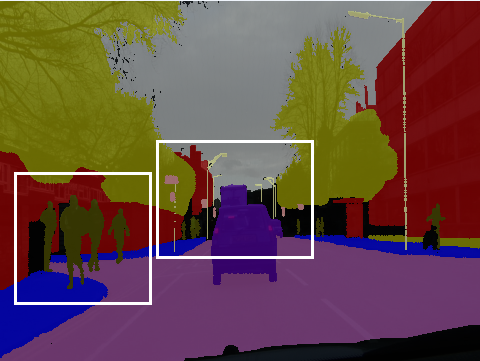}
   \label{fig:test_gt}
}
\end{subfigure}
\begin{subfigure}[F-CDRN-P]{
\includegraphics[width=0.234\textwidth]{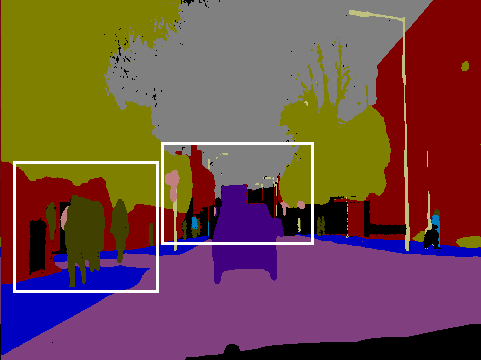}
   \label{fig:test_pools}
}
\end{subfigure}

\begin{subfigure}[FC-DRN-D]{
\includegraphics[width=0.224\textwidth]{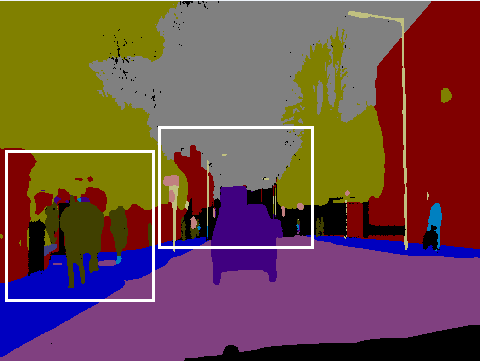}
   \label{fig:test_dils}
}
\end{subfigure}
\begin{subfigure}[FC-DRN-P-D]{
\includegraphics[width=0.224\textwidth]{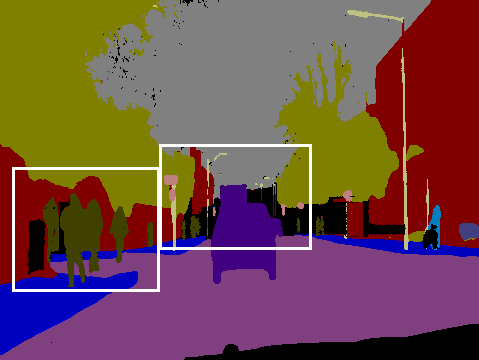}
   \label{fig:test_finetune}
}
\end{subfigure}
\begin{subfigure}[FC-DRN-S]{
\includegraphics[width=0.224\textwidth]{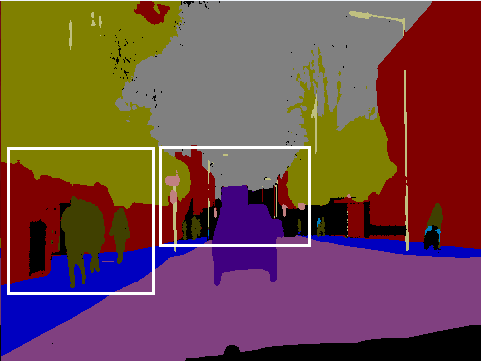}
   \label{fig:test_finetune}
}
\end{subfigure}
\begin{subfigure}[FC-DRN-S-D]{
\includegraphics[width=0.224\textwidth]{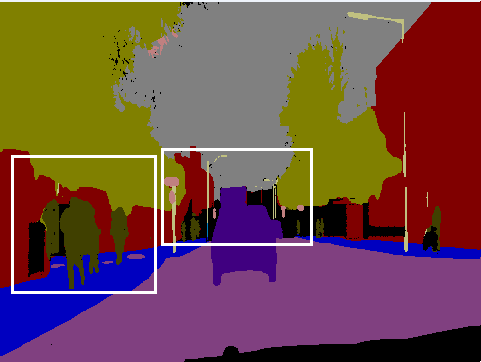}
   \label{fig:test_finetune}
}
\end{subfigure}

\caption{Qualitative results on the test set: (a) input image, (b) ground truth, (c) FC-DRN-P prediction, (d) FC-DRN-D prediction, (e) FC-DRN-P-D prediction, (f) FC-DRN-S prediction and (g) FC-DRN-S-D prediction. Main differences are highlighted with white boxes.}
\label{fig:pred_comparison}
\vspace{-0.1in}
\end{figure*}

Figure~\ref{fig:pred_comparison} presents qualitative results from all architectures. As illustrated in the figure, FC-DRN-P prediction seems to better capture the global information, when compared to FC-DRN-D. This can be observed on the left part of the predictions, where dilated convolutions predict different classes for isolated pixels. Max-poolings understand better the scene and output a cleaner and more consistent prediction. Although FC-DRN-P has a more global view of the scene and is less prone to make local mistakes, it lacks resolution to make fine-grained predictions. In the FC-DRN-P-D prediction, we can see that dilated convolutions help recover the column poles in the middle of the image that were not properly identified in the other architectures, and the predictions of the pedestrians on the left are also sharper. However, the model still preserves the global information of FC-DRN-P, reducing some errors that were present on the left part of the image in the FC-DRN-D prediction. If we take a look at the FC-DRN-S prediction, we can see that it recovers the pedestrian on the right successfully, but fails to correctly segment the sidewalk and the pedestrians on the left. Finetuning this architecture with dilations (FC-DRN-S-D) helps capture a lost pedestrian on the left, but still lacks the ability to sharply sketch the right column pole. Furthermore, there are some artifacts on the top part of the image in both strided convolution-based architectures.

 \begin{table}[tb]
 \footnotesize
   \begin{tabular}{L{2cm} C{2.2cm} C{2.2cm}} 
 \hline
 \textbf{Architecture} & \textbf{mean IoU [\%]} & \textbf{accuracy [\%]}\\
 \hline
 FC-DRN-P & \textbf{81.1} &\textbf{96.1} \\ \hline
 FC-DRN-S & 80.3 &95.9\\ \hline
 FC-DRN-D & 77.4 & 95.5\\ \hline \hline
FC-DRN-P-D& \textbf{81.7} & \textbf{96.0}\\ \hline
  FC-DRN-S-D & 81.1 & 96.0\\ \hline
 
 \end{tabular}
 \caption{Comparison of FC-DRN architectures with different transformations. Results reported on the validation set.}
 
 \label{tab:comparison}
 \vspace{-0.2in}
 \end{table}

\subsection{Results}
Following the comparison in Table \ref{tab:comparison}, we report our final results on the FC-DRN-P-D architecture. Recall that this architecture is a pre-trained FC-DRN with a dense block of 4 max pooling operations (downsampling path) and 4 repeat and convolve operations (upsampling path) and finetuned by substituting the last two max poolings and the first two upsamplings by dilated convolutions, on the same data.

Table \ref{tab:test_results} compares the performance of our model to state-of-the-art models. As shown in the table, our FC-DRN-P-D exhibits state-of-the-art performance when compared to previous methods, especially when it comes to segmenting under-represented classes such as column poles, pedestrians and cyclists. It is worth noting that our architecture improves upon pre-trained models with 10 times more parameters. When compared to FC-DenseNets, FC-DRN outperform both FC-DenseNet67 (with a comparable number of parameters) and 
FC-DenseNet103 (with only $41.5\%$ of its parameters) by $2.5\%$ and $1.4\%$ mean IoU, respectively. Moreover, it also exceeds the performance of more recent methods such as G-FRNet, which uses gated skip connections between encoder and decoder in a FCN architecture, while only using 13\% of its parameters.

In order to further boost the performance of our network, we finetuned it by using soft targets \cite{Hinton2015,Romero15-iclr} ($0.9$ and $0.01$, instead of 1 and 0 in the target representation), improving generalization and obtaining a final score of $69.4\%$ mean IoU and $91.6\%$ global accuracy. Using soft targets allows the network to become more accurate in predicting classes such as pedestrian, fence, column pole, sign, building and sidewalk when compared to the original version. FC-DRN recovers slim objects such as column poles and pedestrians much better than other architectures presented in the literature, while maintaining a good performance on classes composed of larger objects.

It is worth mentioning that, unlike most of current state-of-the-art methods, FC-DRN have not been pre-trained on large datasets such as ImageNet~\cite{ILSVRC15}. Moreover, there are other methods in the literature that exploit virtual images to augment the training data~\cite{Richter} or that leverage temporal information to improve performance~\cite{Jampani2016}. Note that those enhancements complement each other and FC-DRN could most likely benefit from them to boost their final performance as well. However, we leave those as future work.

\begin{figure*}[]
 \centering
\begin{subfigure}{
\includegraphics[width=.65\textwidth]{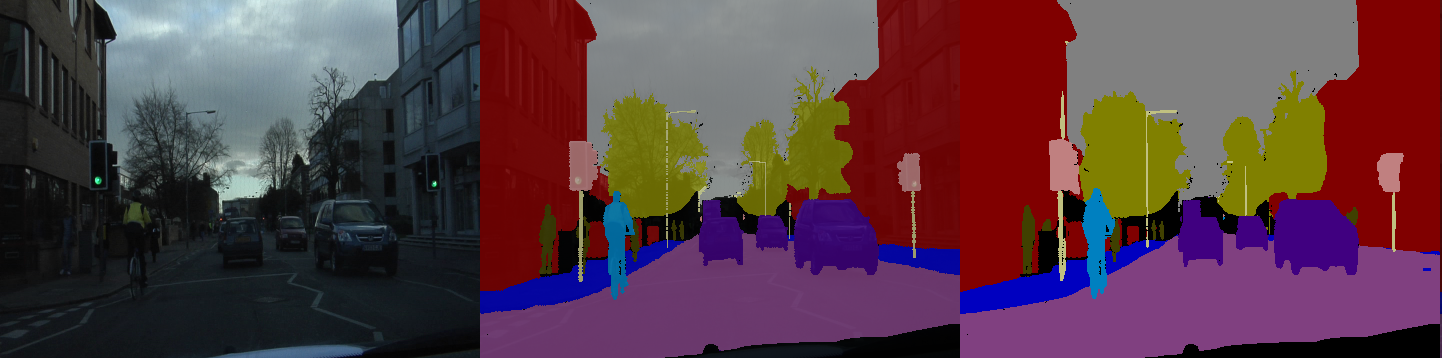}
   \label{fig:test1}
}
\end{subfigure}

\begin{subfigure}{
\includegraphics[width=.65\textwidth]{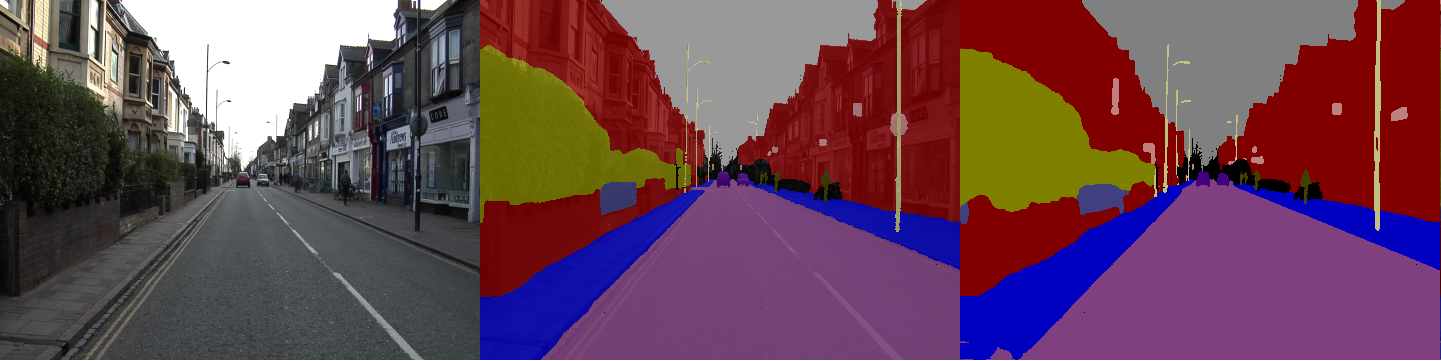}
   \label{fig:test2}
}
\end{subfigure}

\caption{Qualitative results on the CamVid test set. Left: images, middle: ground truths, right: FC-DRN predictions.}
\label{fig:qualitative_results}
\end{figure*}

Figure~\ref{fig:qualitative_results} shows some FC-DRN segmentation maps (right) compared to their respective ground truths (middle). We can observe that the segmentations have good quality, aligned with the quantitative results we obtained. The column poles and pedestrians are sharply segmented, but some difficulties arise when trying to distinguish between sidewalks and roads or in the presence of small road signs.
\begin{table*}[tb]
\footnotesize
\begin{tabular}{L{2.5cm}  C{0.7cm}  C{0.5cm}  C{0.5cm}  C{0.5cm}  C{0.5cm} C{0.5cm} C{0.5cm} C{0.5cm} C{0.5cm} C{0.5cm} C{0.5cm} C{0.5cm} C{0.5cm} | C{0.7cm} C{0.7cm}} 
 \hline
 \textbf{Model} &\textbf{\# params [M]} &\textbf{\rotatebox[origin=c]{90} {Ext. Data}} & \textbf{\rotatebox[origin=c]{90}{Building}} & \textbf{\rotatebox[origin=c]{90}{Tree}} &  \textbf{\rotatebox[origin=c]{90}{Sky}} &  \textbf{\rotatebox[origin=c]{90}{Car}} &  \textbf{\rotatebox[origin=c]{90}{Sign}} &  \textbf{\rotatebox[origin=c]{90}{Road}} &  \textbf{\rotatebox[origin=c]{90}{Pedestrian}} &  \textbf{\rotatebox[origin=c]{90}{Fence}} &  \textbf{\rotatebox[origin=c]{90}{Column pole}} &  \textbf{\rotatebox[origin=c]{90}{Sidewalk}} &  \textbf{\rotatebox[origin=c]{90}{Cyclist}} & \textbf{mean IoU [\%]} & \textbf{Gl. acc. [\%]} \\ \hline
SegNet \cite{SegNet2015} & 29.5 & Yes & 68.7 & 52.0 & 87.0 & 58.5 & 13.4 & 86.2 & 25.3 & 17.9 & 16.0 & 60.5 & 24.8 & 46.4 & 62.5 \\  \hline
DeconvNet \cite{noh2015learning} & 252 & Yes & \multicolumn{11}{c|}{n/a} &  48.9 & 85.9\\ \hline
FCN8 \cite{Long2015} & 134.5 & Yes & 77.8 & 71.0 & 88.7 & 76.1 & 32.7 & 91.2 & 41.7 & 24.4 & 19.9 & 72.7 & 31.0 & 57.0 & 88.0\\  \hline
Visin et al. \cite{Visin2015} & 32.3 & Yes & \multicolumn{11}{c|}{n/a} & 58.8 & 88.7\\ \hline
DeepLab-LFOV~\cite{Chen2014} & $37.3$ & Yes & 81.5 & 74.6 & 89.0 & 82.2 & 42.3 & 92.2 & 48.4  & 27.2 & 14.3 & 75.4 & 50.1 & 61.6 & -\\ \hline
Bayesian SegNet \cite{KendallBC15} & 29.5 & Yes& \multicolumn{11}{c|}{n/a} &  63.1 & 86.9\\ \hline
Dilation8  \cite{Yu2015} & 140.8 & Yes & 82.6 & 76.2 & 89.0 & 84.0 & 46.9 & 92.2 & 56.3 & 35.8 & 23.4 & 75.3 & 55.5 & 65.3 & 79.0 \\ \hline
FC-DenseNet67~\cite{Jegou17}  &3.5 & No & 80.2 & 75.4 & 93.0 & 78.2 & 40.9 & \textbf{94.7} & 58.4 & 30.7 & 38.4 & 81.9 & 52.1 & 65.8 & 90.8\\ \hline
Dilation8 + FSO~\cite{kundu2016feature}&  130 & Yes & \textbf{84.0} & 77.2 & 91.3 & \textbf{85.6} & \textbf{49.9} & 92.5 & 59.1 & 37.6 & 16.9 & 76.0 & 57.2 & 66.1 &88.3\\ \hline
FC-DenseNet103~\cite{Jegou17} & 9.4 & No & 83.0 & \textbf{77.3} & \textbf{93.0 }& 77.3 & 43.9 & 94.5 & 59.6 & 37.1 & 37.8 & 82.2& 50.5 & 66.9 &91.5 \\ \hline
G-FRNet~\cite{islam2017gated} & 30 & Yes & 82.5 & 76.8 & 92.1 & 81.8 & 43.0 & 94.5 & 54.6 & \textbf{47.1} & 33.4 &\textbf{82.3}& 59.4 & 68.0 &90.8\\ \hline \hline
FC-DRN -P-D & 3.9 & No & 82.6 & 75.7& 92.6 & 79.9 & 42.3 &94.1 & \underline{61.2} &  36.9 & \underline{42.6} &81.2 & \underline{\textbf{61.8}}& \underline{68.3} &91.4 \\ \hline
FC-DRN-P-D + ST & 3.9 & No & 83.5 & 75.6 & 92.1 & 78.5 & 46.6 & 93.9  & \textbf{62.7} &44.3 & \textbf{43.1} & 82.2 & 60.8 & \textbf{69.4}  &\textbf{91.6} \\ \hline
 \end{tabular}
  \caption{Results on CamVid test set, reported as IoU per class, mean IoU and global accuracy, compared to state-of-the-art.}
  \label{tab:test_results}
  \vspace{-0.2in}
 \end{table*}
 
\section{Delving deeper into FC-DRN transformations}

\begin{figure}[tb]
\begin{subfigure}[Initial FC-DRN architectures]{
\includegraphics[width=0.75\columnwidth]{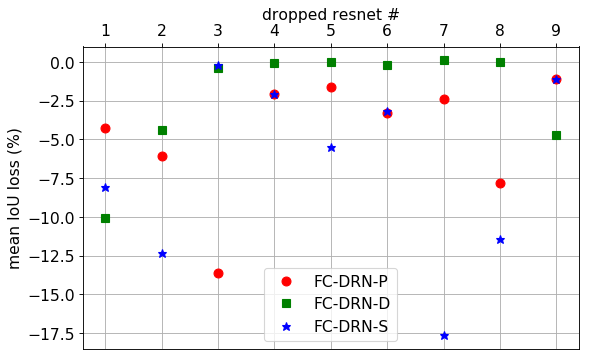}
   \label{fig:dropres_init}
}
\end{subfigure}
 \centering
\begin{subfigure}[Finetuned architectures]{
\includegraphics[width=0.75\columnwidth]{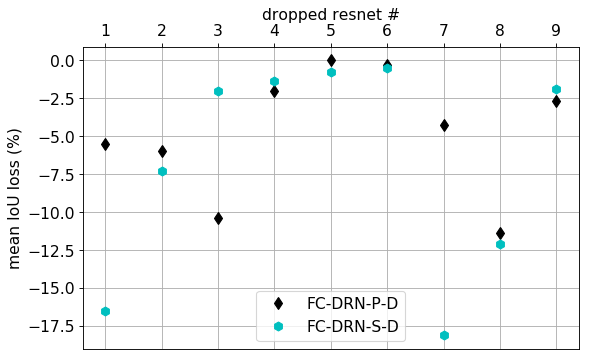}
   \label{fig:dropres_ft}
}\end{subfigure}%
\caption{Results of dropping ResNets from a trained FC-DRN reported for the validation set: y-axis represents the loss in mean IoU when comparing to the model with all ResNets, x-axis represents the ID of the dropped ResNet.}
\label{fig:drop_resnets}
\vspace{-0.2in}
\end{figure}

\begin{figure*}[]
\begin{subfigure}[Initial FC-DRN architectures]{
\includegraphics[width=1.0\textwidth]{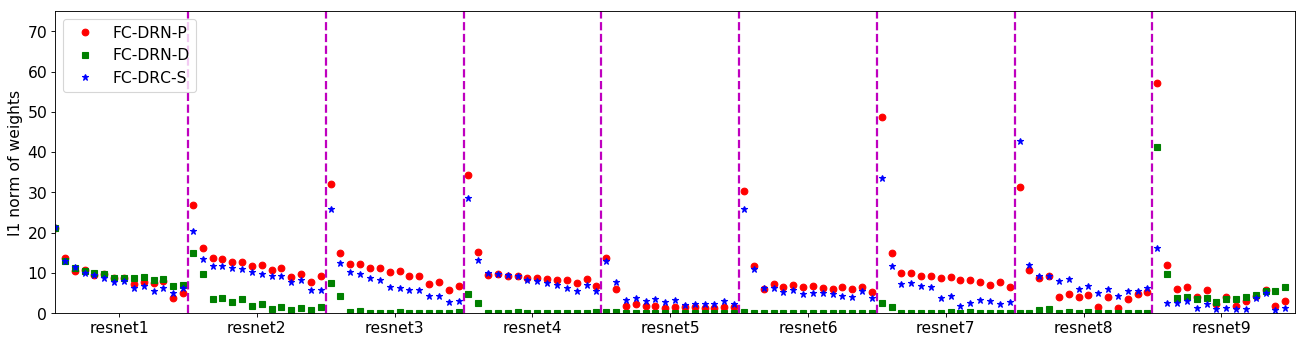}
   \label{fig:weights_init}
}
\end{subfigure}
 \centering
\begin{subfigure}[Finetuned architectures]{
\includegraphics[width=1.0\textwidth]{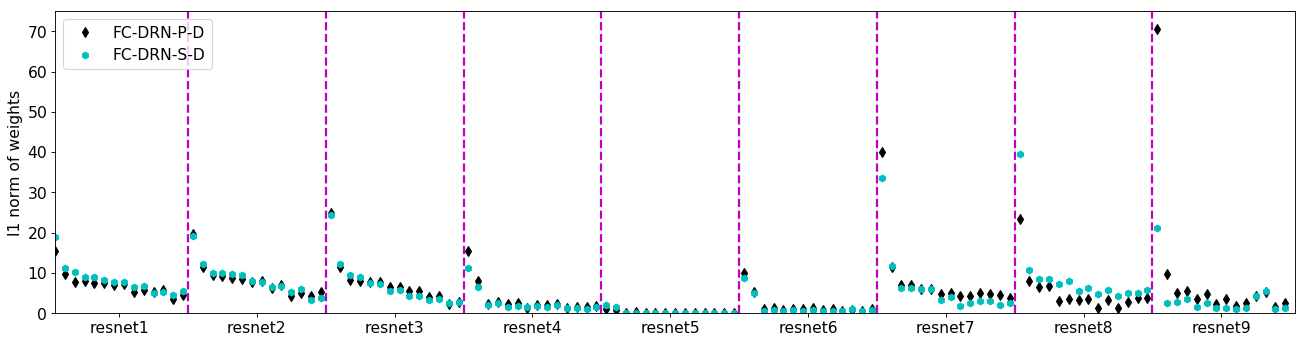}
   \label{fig:weights_ft}
}\end{subfigure}%
\caption{Norms of the weights in ResNets from trained FC-DRN variants. Different ResNets separated with vertical lines.}
\label{fig:weights}
\vspace{-0.2in}
\end{figure*}

In this section, we provide an in depth analysis of the variants of the trained FC-DRN architectures to compare different transformation operations: pooling, dilation and strided convolution. We start by dropping ResNets from a FC-DRN and then look into the weight's norms of all ResNets in the models. We end the section with a discussion exploiting the observations from the network inspection.

We follow the strategy of dropping layers introduced in \cite{Veit2016,Huang2016_stodepth}, and drop all residual blocks of a ResNet (we only keep the first residual block that adjusts the depth of the feature maps) with the goal of analyzing the implications of using different transformation operations. The results of the experiment are shown in Figure \ref{fig:drop_resnets}. On one hand, Figure \ref{fig:dropres_init} reports the performance drops in percentage of mean IoU for each ResNet in the initial networks (i.e. FC-DRN-P, FC-DRN-D and FC-DRN-S). Surprisingly, dropping ResNets 3 to 8 barely affects the performance of FC-DRN-D. However, both pooling and strided convolution models suffer from the loss of almost any ResNet. On the other hand, Figure \ref{fig:dropres_ft} presents the results of dropping ResNets in the finetuned models (i.e. FC-DRN-P-D and FC-DRN-P-S). Finetuning the pooling network with dilations makes ResNets 5 and 6 slightly less necessary, while ResNet 8 becomes the most relevant one. Finetuning the strided convolution network with dilations makes ResNet 1 extremely important, while ResNets 4 to 6 have a smaller influence. In general, it seems that finetuning with dilations reduces the importance of the bottleneck ResNets of the network. Moreover, the results might suggest that dilations do not change the representation level as much as poolings/strided convolutions.

To gain further insight on what FC-DRN variants are learning, we visualize the $\ell_1$ norm of the weights in all ResNets. More precisely, given a $4D$ weight tensor $w$ of a shape $[N, M, K, K]$ that applies the transformation between two consecutive layers with $N$ and $M$ channels, respectively, we compute: $1/M\sum_m ||w[:,m,:,:]||_1$. The results of this experiment for different FC-DRN variants are shown in Figure \ref{fig:weights}. The weight norms shown in the figure are in line with the discussion of Figure \ref{fig:drop_resnets}. On one hand, basic FC-DRN architectures are displayed in Figure \ref{fig:weights_init}. For FC-DRN-P, we can see how the weights in ResNet 5 have lower values, suggesting an explanation for the lower drop in performance when removing ResNet 5. Furthermore, we can also see that the weight norms in the case of FC-DRN-D are almost zero for ResNets 3 to 8, suggesting that the network does not fully exploit its capacity. In the case of FC-DRN-S, the norms of the weights in ResNet 5 are smaller (similar to FC-DRN-P), whereas ResNet 2 exhibits some of the highest norms. On the other hand, Figure \ref{fig:weights_ft} shows the results of finetuning. FC-DRN-P-D has weight norms among ResNets 4 to 6 lower than FC-DRN-P, whereas FC-DRN-S-D follows a similar pattern when compared to FC-DRN-S. However, from the observed weights, it would seem as ResNet 4 still benefits from some refinement steps. Overall, it seems that finetuning consistently reduces the weight norms of the bottleneck ResNets.

It is important to note that the structure of ResNets, due to the usage of the residual block, allows the model to self-adjust its capacity when needed, forcing the residual transformation of the residual block to be close to $0$ and using the identity connection to forward the information. We hypothesize that this behavior of residuals is observed for some layers in our model (as it is shown in Figure \ref{fig:weights}). To test our hypothesis, we decided to reduce the capacity of \emph{trained} FC-DRN by removing layers from the residuals of ResNets for which the norm of weights is small and to monitor the performance of the \emph{compressed} FC-DRN model. If our hypothesis is true, then removing the residuals in the layers where the norm is close to $0$ should not affect strongly the model performance. We choose to drop the layers whose weight norms are close enough to zero, based on visual inspection of Figure \ref{fig:weights}, thus allowing each representation level to have different number of refinement steps. The results of this trained model compression experiment are reported in Table \ref{tab:compression}. We can see that after removing $8\%$ of the parameters from FC-DRN-P and FC-DRN-S, there is a drop in mean IoU on validation set of $-1.6$ and $-5.4$, respectively. Interestingly, we were able to remove $38\%$ of weights from FC-DRN-D model while only experiencing a drop of $-0.8$ in mean IoU. Both finetuned models can be compressed by removing $15\%$ of the capacity with slight performance drops of $-1$ and $-1.7$ for FC-DRN-P-D and FC-DRN-S-D, respectively. In general, it seems that finetuning the models with dilations not only improves the segmentation results but also makes the model more compressible.

Finally, we test if the optimization process of the high capacity FC-DRN reaches better local minima, due to self-adjustment of ResNets' capacity, than if we train a low capacity FC-DRN from scratch. To this end, we trained from scratch the reduced capacity FC-DRN-D model and compared the results to the numbers reported in Table \ref{tab:compression}. The re-trained model obtained the mean validation IoU of $76.6$. This is $0.8\%$ below the result reported for the high capacity FC-DRN-D. We hypothesize that the model capacity reduction during the optimization process helps in reaching a better local minima. Since the self-adjustment of the ResNet capacity might be encouraged by weight decay, for the sake of the comparison we also trained the reduced capacity model without weight decay at all. This model obtained the mean validation IoU of $75.4\%$ that is $2\%$ below the full capacity model.

 \begin{table}[tb]
 \footnotesize
   \begin{tabular}{L{2cm} C{2.2cm} C{2.2cm}} 
 \hline
 \textbf{Architecture} & \textbf{mean IoU loss [\%]} & \textbf{compression rate}\\
 \hline
 FC-DRN-P &  $-1.6$ & $1.08$ \\ \hline
 FC-DRN-D &  $-0.8$ & $1.38$ \\ \hline
 FC-DRN-S &  $-5.4$ & $1.08$ \\ \hline \hline
 FC-DRN-P-D & $-1$ & $1.15$ \\ \hline
 FC-DRN-S-D & $-1.7$ & $1.15$ \\ \hline
 \end{tabular}
 \caption{Performance drop for different models when removing all layers with small norm of weights from a trained FC-DRN variant. Compression rate represents the ratio between the number of parameters in the original model and the number of parameters after removing layers. Results reported on the validation set.}
 \label{tab:compression}
 \vspace{-0.2in}
 \end{table}

\section{Conclusions}
In this paper, we combined two standard image segmentation architectures (FC-DenseNets and FC-ResNets) into a single network that we called Fully Convolutional DenseResNet. Our FC-DRN fuses the benefits of both models: gradient flow and iterative refinement from FC-ResNets as well as multi-scale feature representation and deep supervision from FC-DenseNets. We demonstrated the potential of our model on the challenging CamVid urban scene understanding benchmark and reported state-of-the-art results, with at least $2$x fewer parameters than concurrent models.

Additionally, we analyzed different downsampling operations used in the context of semantic segmentation: dilated convolution, strided convolution and pooling. We inspected the FC-DRN by dropping ResNets from the trained models as well as by visualizing the weight norms of different layers and showed that ResNets (by model construction) are good regularizers, since they can reduce the model capacity when needed. In this direction, we observed that coarser representations seem to benefit from less refinement steps. Moreover, our results comparing different transformations suggest that pooling offers the best generalization capabilities, while the benefits of dilated convolutions only apply when combined with pre-trained networks that contain downsampling operations. 

\subsubsection*{Acknowledgments}
The authors would like to thank the developers of Pytorch \cite{pytorch}. We acknowledge the support of the following agencies for research funding and computing support: CIFAR, Canada Research Chairs, Compute Canada and Calcul Qu\'{e}bec, as well as NVIDIA for the generous GPU support.

{\small
\bibliographystyle{ieee}
\bibliography{egbib}
}

\clearpage

\appendix
\section{Supplementary Material}
\label{app:arch}
We present the architecture details in Table~\ref{tab:arch_details}, agnostic to the type of transformation used in between ResNets. 

The outputs of transformation blocks are reused when needed. In the case of dilations, we still maintain all transformations to keep the number of parameters roughly constant. The right column in the table indicates the number of feature channels after applying each operation.

\begin{figure}[h]
 \centering
\begin{subfigure}[ResNet basic block]{
\includegraphics[width=0.28\columnwidth, angle =90]{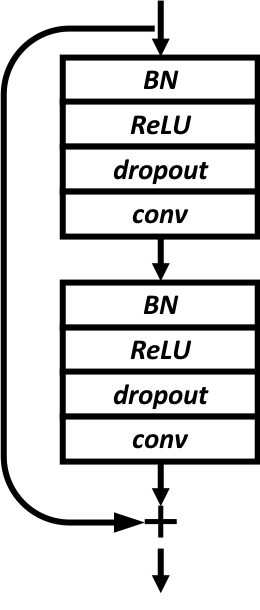}
   \label{subfig:bb}
}\end{subfigure}\hspace{0.07\columnwidth}
\centering
\begin{subfigure}[Multi-grid block]{
\includegraphics[width=0.28\columnwidth, angle =90]{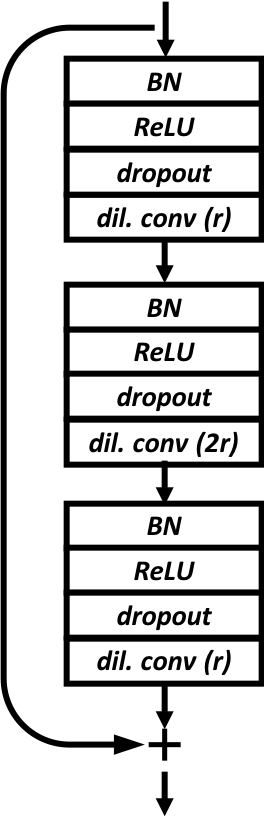}
   \label{subfig:mg}
}
\end{subfigure}

\caption{ResNet block and multi-grid dilation block used in our architecture. For multi-grid dilation block, we use $r$ to represent dilation factor.}
\label{fig:blocks}
\end{figure}

The detailed composition of the ResNet block and the multi-grid dilation block used in our architecture is presented in Figure~\ref{fig:blocks}.

\begin{figure*}[!t]
 \centering
\begin{subfigure}{
\includegraphics[width=.9\textwidth]{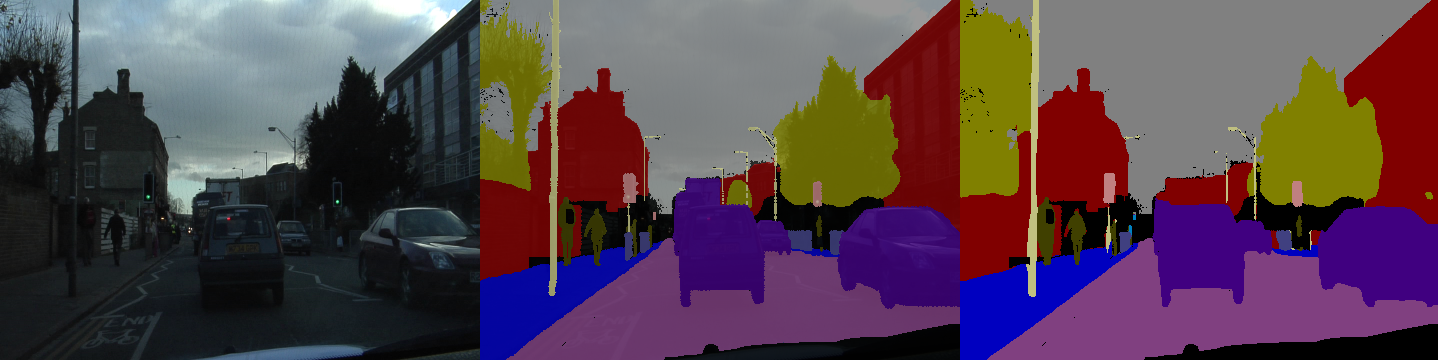}
   \label{fig:test_pools}
}
\end{subfigure}
\begin{subfigure}{
\includegraphics[width=.9\textwidth]{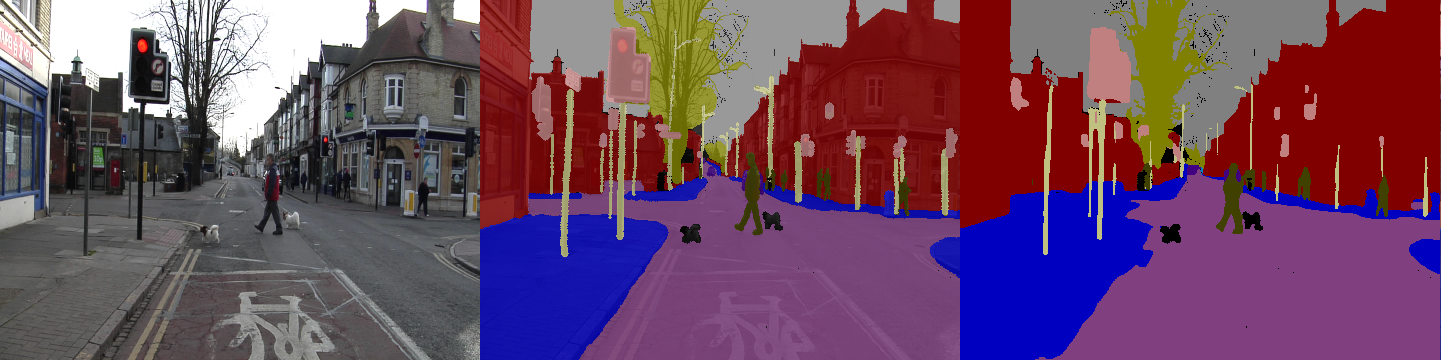}
   \label{fig:test3}
}
\end{subfigure}
\caption{Additional qualitative results on CamVid test set. Test images are shown on the left, ground truth on the middle and FC-DRN predictions on the right.}
\label{fig:qualitative_results_app}
\end{figure*}

Additionally, we also present some additional output segmentations for FC-DRN-P-D, the max-pooling architecture finetuned with some dilated convolutions and trained with soft targets. Predictions are shown in Figure~\ref{fig:qualitative_results_app}.

\vspace{1in}

\newcolumntype{P}[1]{>{\centering\arraybackslash}p{#1}}
\begin{table}[h!]
\footnotesize
\begin{tabular}{ | P{7cm} P{0.5cm}|}
 \hline
 Operation & Out\\
 \hline\hline
 IDB: $3\times3$ conv , max pool, 2 $3\times3$ conv & 50 \\
 \hline
 \textbf{R1} & 30 \\
 \hline
  [${TF_d}$ (IDB), ${TF_d}$ (R1)] & 80 \\
  \hline
 mixing block & 80\\
 \hline
  \textbf{R2} & 40 \\
     \hline
   [${TF_d}^2$ (IDB), ${TF_d}^2$ (R1),${TF_d}$ (R2) ] & 120 \\
  \hline
   mixing block & 120\\
 \hline
   \textbf{R3} & 40 \\
     \hline
   [${TF_d}^3$ (IDB), ${TF_d}^3$ (R1), ${TF_d}^2$ (R2), ${TF_d}$ (R3) ] & 160 \\
  \hline
   mixing block & 160\\
 \hline
   \textbf{R4} & 40 \\
     \hline
   [${TF_d}^4$ (IDB), ${TF_d}^4$ (R1), ${TF_d}^3$ (R2), ${TF_d}^2$ (R3),  ${TF_d}$ (R4) ] & 200 \\
  \hline
   mixing block & 200\\
 \hline
   \textbf{R5} & 50 \\
     \hline
   [${TF_d}^3$ (IDB), ${TF_d}^3$ (R1), ${TF_d}^2$ (R2), ${TF_d}$ (R3), R4, ${TF_u}$ (R5) ] & 200 \\
  \hline
   mixing block & 200\\
 \hline
   \textbf{R6} & 40 \\
     \hline
   [${TF_d}^2$ (IDB), ${TF_d}^2$ (R1), ${TF_d}$ (R2), R3,  ${TF_u}$ (R4), ${TF_u}^2$ (R5), ${TF_u}$ (R6) ] & 240 \\
  \hline
   mixing block & 240\\
 \hline
   \textbf{R7} & 40 \\
     \hline
   [${TF_d}$ (IDB), ${TF_d}$ (R1), R2, ${TF_u}$ (R3),  ${TF_u}^2$ (R4), ${TF_u}^3$ (R5), ${TF_u}^2$ (R6), ${TF_u}$ (R7) ] & 280 \\
  \hline
   mixing block & 280\\
 \hline
   \textbf{R8} & 40 \\
     \hline
   [IDB, R1, ${TF_u}$ (R2), ${TF_u}^2$ (R3),  ${TF_u}^3$ (R4), ${TF_u}^4$ (R5),  ${TF_u}^3$ (R6), ${TF_u}^2$ (R7),  ${TF_u}$ (R8) ] & 320 \\
  \hline
   mixing block & 320\\
 \hline
   \textbf{R9} & 30 \\
   \hline
   [IDB, R1, ${TF_u}$ (R2), ${TF_u}^2$ (R3),  ${TF_u}^3$ (R4), ${TF_u}^4$ (R5),  ${TF_u}^3$ (R6), ${TF_u}^2$ (R7),  ${TF_u}$ (R8), R9 ] & 350 \\
     \hline
 mixing block & 350\\
 \hline
  FUB: 2x2 repeat upsampling, $3\times3$ conv & 50\\
 \hline
  Linear classifier: $1\times1$ conv & 11\\
 \hline
\end{tabular}
\label{tab:arch_details}
\caption{Architecture details. ResNets are indicated as \textit{R}, the initial downsampling block as \textit{IDB} and the final upsampling Block as \textit{FUB}. We use ${TF_d}^x$ and ${TF_u}^i$ to denote transformation blocks in the downsampling path and in the upsampling path, respectively. The superscript $i$ is the number of cascaded transformations applied to their input.} \label{tab:arch_details}
\end{table}

\end{document}